\documentstyle[amsfonts,epsf,amsbsy,chicago,twocolumn]{article}

\newcommand{\BiInfinity}			{ \stackrel{\leftrightarrow} {S} }

\newcommand{\Past}        		{ \stackrel{\leftarrow} {S} }
\newcommand{\past}        		{ {\stackrel{\leftarrow} {s}} }
\newcommand{\Future}      		{ \stackrel{\rightarrow}{S} }
\newcommand{\future}      		{ \stackrel{\rightarrow}{s} }

\newcommand{\PastLt}			{ {\stackrel{\leftarrow} {S}}_t^L }

\newcommand{\FutureL}			{ {\stackrel{\rightarrow}{S}}^L }
\newcommand{\FutureLt}			{ {\stackrel{\rightarrow}{S}}_t^L }

\newcommand{\AllPasts}			{ { \stackrel{\leftarrow} {\rm {\bf S}} } }

\newcommand{\CausalState}		{ {\cal S} }
\newcommand{\causalstate}		{ \sigma }
\newcommand{\CausalStateSet}		{ \boldsymbol{\CausalState} }
\newcommand{\AlternateState}		{ {\cal R} }
\newcommand{\alternatestate}		{ \rho }
\newcommand{\AlternateStateSet}	{ \boldsymbol{\AlternateState} }
\newcommand{\PrescientState}		{ \hat{\AlternateState} }
\newcommand{\prescientstate}		{ \hat{\alternatestate} }
\newcommand{\PrescientStateSet}	{ \boldsymbol{\PrescientState}}
\newcommand{\CausalEquivalence}	{ {\sim}_{\epsilon} }

\newcommand{\Prob}				{ {\rm P} }
\newcommand{\ProbAnd}			{ {,\;} }

\newcommand{\Cmu}				{ C_\mu }

\newcommand{\EE}				{ {\bf E}}
\newtheorem{theorem}{Theorem}
\newtheorem{lemma}{Lemma}

\newtheorem{definition}{Definition}

\begin{document}

\bibliographystyle{chicago}

% *******************Title Page************************
\title{Pattern Discovery and Computational Mechanics}
\author{Cosma Rohilla Shalizi\thanks{Permanent address: Physics
Department, University of Wisconsin, Madison, WI 53706} \\
and James P. Crutchfield \\
{\it Santa Fe Institute, 1399 Hyde Park Road, Santa Fe, NM 87501}\\
Electronic addresses: \{shalizi,chaos\}@santafe.edu}
\date{27 January 2000}
\maketitle

%*************************Abstract*********************
\begin{abstract}

Computational mechanics is a method for discovering, describing and quantifying
patterns, using tools from statistical physics. It constructs optimal,
minimal models of stochastic processes and their underlying causal structures.
These models tell us about the intrinsic computation embedded within a
process---how it stores and transforms information. Here we summarize the
mathematics of computational mechanics, especially recent optimality and
uniqueness results. We also expound the principles and motivations underlying
computational mechanics, emphasizing its connections to the minimum
description length principle, PAC theory, and other aspects of machine learning.

\begin{center}
Santa Fe Institute Working Paper 00-01-008

{\bf Keywords}: pattern discovery, machine learning, computational mechanics,
information, induction, $\epsilon$-machine
\end{center}

\end{abstract}
%******************************************************

\tableofcontents

%************************* INTRODUCTION *************************

\section{Introduction}

All students of machine learning are familiar with pattern recognition; in this
paper we wish to introduce a new term for a related, relatively
under-recognized concept, {\it pattern discovery}, and a way of tackling such
problems, {\it computational mechanics}.

The term pattern {\it discovery} is meant to contrast with both pattern {\it
recognition} and pattern {\it learning}.  In pattern recognition, the goal of
the system is to accurately assign inputs to pre-set categories.  In most
learning systems, the goal is to determine which of several pre-set
categorization schemes is correct.  (Naturally, the two tasks are closely
connected \cite{Vapnik-nature}.)  In either case, the representations used have
been, as it were, handed down from on high, due to choices external to the
recognition and learning procedures.

In pattern {\it discovery}, however, the aim is to avoid, so far as possible,
such {\it a priori} assumptions about what structures are relevant.  This is,
of course, an ancient problem, and one which has not been ignored in machine
learning. While there are ingenious schemes for pattern discovery via trial and
error, some even informed by empirical psychology \cite{Holland-induction}, we
believe that a more direct approach is not only possible but also illuminates
the ideal results and the limitations of all pattern-discovery methods.

Computational mechanics originated in physics as a complementary approach to
statistical mechanics for dealing with complex, organized systems
\cite{Calculi-of-emergence}.  In such systems the ``forward'' approach of
statistical mechanics---deriving macroscopic properties from the interactions
of microscopic components---is often intractable, though data can be had in
abundance.\footnote{But see \citeN{Chaikin-Lubensky} and
\citeN{Cross-Hohenberg} for organized systems where the ``forward'' approach
works.}  Computational mechanics follows an ``inverse'' strategy, extending the
idea of extracting ``geometry from a time series''
\cite{Geometry-from-a-time-series}.  It builds the simplest model capable of
capturing the patterns in the data---a representation of the causal structure
of the hidden process which generated the observed behavior.\footnote{See
\citeN{DNCO} for an example of using both statistical and computational
mechanics to analyze the same physical system.}  In a sense that will be made
clear as we go on, this representation---the $\epsilon$-machine---is the unique
maximally efficient model of the observed data-generating process.  The basic
ideas of computational mechanics were introduced in
\citeN{Inferring-stat-compl}.  Since then they have been used to analyze
dynamical systems, cellular automata, hidden Markov models, evolved spatial
computation, stochastic resonance, globally coupled maps, and the dripping
faucet experiment; see \citeN[Sec. 1]{CMPPSS} for references.

This paper is arranged as follows.  First we examine some conceptual issues
about pattern discovery and the way they are addressed by computational
mechanics.  We devote the bulk of this paper to a summary of the mathematical
structure of computational mechanics, with particular attention to optimality
and uniqueness theorems.  Results are stated without proof; readers will find a
full treatment in \citeN{CMPPSS}.  Then we discuss the ties between
computational mechanics and several approaches to machine learning.  Finally,
we close by pointing out directions for future theoretical work.

\section{Conceptual Issues}

Any approach to handling patterns should, we claim, meet a number of
criteria; the justifications for which are given in \citeN{CMPPSS} in
detail. It should be at once

\begin{enumerate}
\label{Desiderata}

\item {\em Predictive}, i.e., the models it produces should allow us to predict
the original process or system we are trying to understand; and, by that
token, provide a compressed description of it;

\item {\em Computational}, showing how the process stores, transmits, and
transforms information;

\item {\em Calculable}, analytically or by systematic approximation;

\item {\em Causal}, telling us how instances of the pattern are actually
produced; and

\item {\em Naturally stochastic}, not merely tolerant of noise but explicitly
formulated in terms of ensembles.

\end{enumerate}

In any modeling approach, the two (related) problems are to devise a mapping
from states of the world (or, more modestly, states of inputs) to states of the
model, and to accurately and precisely predict future states of the world on
the basis of the evolution of the model.  (Cf.~\citeN{Holland-induction} on
``q-morphisms''.)  The key idea of computational mechanics is that the
information required to do this is actually {\it in} the data, provided there
is enough of it.  In fact, if we go about it right, the key step is getting the
mapping from data to model states right---equivalently, the problem is to
decide which data-sets should be treated as equivalent and how data should be
partitioned.
Once we have the correct mapping of data into equivalence classes, accurate
prediction is actually fairly simple.  That the correct mapping should treat as
equivalent all data-sets which leave us in the same degree of knowledge about
the future has a certain intuitive plausibility, but also sounds hopelessly
vague.  In fact, we can specify such a partition in a precise, operational way,
show that it is the best one to use, and determine it empirically.  We call the
function which induces that partition $\epsilon$, and its equivalence classes
{\it causal states}.  In fact, the model we get from using such a partition ---
the $\epsilon$-machine --- meets all the criteria stated above.  It is
because the $\epsilon$-machine shows, in a
very direct way, how information is stored in the process, and how that stored
information is transformed by new inputs and by the passage of time,
that computational mechanics is about {\it computation}.

\section{Mathematical Development}

\subsection{Note on Information Theory}

The bulk of the following development will be consumed with notions and results
from information theory.  We follow the standard definitions and notation of
\citeN{Cover-and-Thomas}, to which we refer readers unfamiliar with the
theory.  In particular, $H[X]$ is the entropy of the discrete random variable
$X$, interpreted as the uncertainty in $X$, measured in bits.\footnote{Here,
and throughout, we follow the convention of using capital letters to denote
random variables and lower-case letters their particular values.}  $H[X|Y]$ is
the entropy of $X$ conditional on $Y$, and $I[X;Y]$ the mutual information
between the two random variables.

\subsection{Hidden Processes}
\label{HiddenProcesses}

We restrict ourselves to discrete-valued, discrete-time stationary stochastic
processes.  (See Sec.~\ref{ThingsThatAreNotYetTheorems} for discussion of these
restrictions.)  Intuitively, such processes are sequences of random variables
$S_i$, the values of which are drawn from a countable set ${\cal
A}$.  We let $i$ range over all the integers, and so get a bi-infinite
sequence $\BiInfinity = \ldots S_{-1} S_0 S_1 \ldots$. In fact, we define a
{\em process} in terms of the distribution of such sequences.

Given that $\BiInfinity$ is well-defined, there are probability distributions
for sequences of every finite length.  Let $\FutureLt$ be the sequence of $S_t,
S_{t+1}, \ldots, S_{t+L-1}$ of $L$ random variables beginning at $S_t$.
$\Future_t^0 \equiv \lambda$, the null sequence.  Likewise, $\PastLt$ denotes
the sequence of $L$ random variables going up to $S_t$, but not including it:
$\PastLt = \Future_{t-L}^L$.  Both $\FutureLt$ and $\PastLt$ take values from
$s^L \in {\cal A}^L$.  Similarly, $\Future_t$ and $\Past_t$ are the
semi-infinite sequences starting from and stopping at $t$ and taking values
$\future$ and $\past$, respectively.

Requiring the process $S_i$ to be stationary means that
\begin{equation}
\Prob(\FutureLt = s^L) = \Prob(\Future_0^L = s^L) ~,
\end{equation}
for all $t \in {\Bbb Z}$, $L \in {\Bbb Z}^{+}$, and all $s^L \in {\cal A}^L$.
(A stationary process is one that is time-translation invariant.)
Consequently, $\Prob(\Future_t = \future) = \Prob(\Future_0 = \future)$ and
$\Prob(\Past_t = \past) = \Prob(\Past_0 = \past)$, and so the subscripts may be
dropped.

\subsection{Effective States}
\label{ThePool}

\begin{figure}
\epsfxsize=2.7in
\begin{center}
\leavevmode
\epsffile{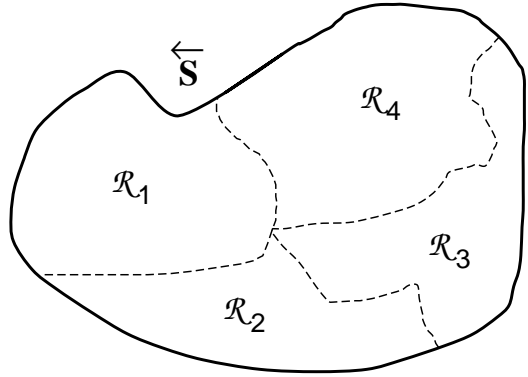}
\end{center}
\caption{A schematic picture of a partition of the set $\AllPasts$
  of all histories into some class of effective states:
  $\AlternateStateSet = \{ {\cal R}_i : i = 1, 2, 3, 4 \}$.  Note that the
  ${\cal R}_i$ need not form compact sets; we simply draw them
  that way for clarity. One should have in mind Cantor sets or other more
  pathological structures.
%  (From \cite{TDCS}, with permission.)
  }
\label{OccamsPool}
\end{figure}

Our goal is to predict all or part of $\Future$ using some function of some
part of $\Past$. We begin by taking the set $\AllPasts$ of all pasts and
partitioning it into mutually exclusive and jointly comprehensive subsets.
That is, we make a class $\AlternateStateSet$ of subsets of pasts.  (See
Fig.~\ref{OccamsPool}.)  Each $\alternatestate \in
\AlternateStateSet$ will be called a {\it state} or an {\it effective state}.
When the current history $\past$ is included in the set $\alternatestate$, we
will say the process is in state $\alternatestate$.  Thus, there is a function
from histories to effective states:
\begin{equation}
\eta: \AllPasts \mapsto \AlternateStateSet ~.
\label{EtaDefn}
\end{equation}
An individual history $\past \in \AllPasts$ maps to a specific state
$\alternatestate \in \AlternateStateSet$; the random variable $\Past$ for the
past maps to the random variable $\AlternateState$ for the effective states.

Any function defined on $\AllPasts$ will serve to partition that set: we just
assign to the same $\alternatestate$ all the histories $\past$ on which the
function takes the same value.  (Similarly, any equivalence relation on
$\AllPasts$ partitions it.)  Each effective state has a well-defined
conditional distribution of futures, though not necessarily a unique one.
Specifying the effective state thus amounts to making a prediction about the
process's future.  In this way, the framework formally incorporates traditional
methods of time-series analysis.

\subsection{Patterns in Ensembles}
\label{PatternsInEnsembles}

It will be convenient to have a way of talking about the uncertainty of the
future.  We do not want to use $H[\Future]$, since that is infinite in general.
Instead, we will work with $H[\FutureL]$, the uncertainty of the next $L$
symbols, treated as a function of $L$.

\begin{definition}[Capturing a Pattern]
\label{capturing-a-pattern}
$\AlternateStateSet$ {\it captures a pattern} iff there exists an
$L$ such that
\begin{equation}
H[\FutureL | \AlternateState] < LH[S] ~.
\end{equation}
\end{definition}
$\AlternateStateSet$ captures a pattern when it tells us something about how
the distinguishable parts of a process affect each other: $\AlternateStateSet$
exhibits their dependence.  (We also speak of $\eta$ as capturing a pattern.)
The smaller $H[\FutureL | \AlternateState]$, the stronger the pattern captured
by $\AlternateStateSet$. Our first result bounds how strongly
$\AlternateStateSet$ can capture a process's pattern.

\begin{lemma}
For all $\AlternateStateSet$ and for all $L \in {\Bbb Z}^{+}$,
\begin{equation}
H[\FutureL|\AlternateState] \geq H[\FutureL|\Past]  ~.
\label{cant-beat-the-past}
\end{equation}
\label{old-country-lemma}
\end{lemma}

\subsection[Minimality and Prediction]{Minimality and Prediction}

Let's invoke Occam's Razor: ``It is vain to do with more what can be done with
less''.  To use the razor, we have to fix what is to be ``done'' and what
``more'' and ``less'' mean.  The job we want done is accurate prediction; i.e.,
to reduce the conditional entropies $H[\FutureL|\AlternateState]$ as far as
possible, down to the bound set by Lemma \ref{old-country-lemma}.  But we want
to do this as simply as possible, with as few resources as possible.  To meet
both constraints---minimal uncertainty and minimal resources---we will need a
measure of the second.  Since $\Prob(\Past = \past)$ is well defined, it
induces a probability distribution on the $\eta$-states, and we can set up the
following measure of resources.

\begin{definition}
{\bf (Complexity of State Classes)}
The statistical complexity of a class $\AlternateStateSet$ of states is
\begin{eqnarray}
\Cmu(\AlternateStateSet) & \equiv & H[\AlternateState] \\ \nonumber
  & = & - \sum_{\alternatestate \in \AlternateState} \Prob(\AlternateState=\alternatestate) \log_2 \Prob(\AlternateState=\alternatestate) ~,
\end{eqnarray}
when the sum converges to a finite value.
\label{statistical-complexity-defined}
\end{definition}

$\Cmu(\AlternateStateSet)$ is the average uncertainty in the process's current
state $\AlternateState$. This is the same as the average amount of memory (in
bits) that the process {\em appears} to retain about the past, given the chosen
state class $\AlternateStateSet$.  We wish to do with as little of this memory
as possible. Our objective, then, is to find a state class which minimizes
$\Cmu$, subject to the constraint of maximally accurate prediction.

\subsection{Causal States}
\label{DefnCausalStatesEMs}

\begin{definition}
{\bf (A Process's Causal States)}
The {\rm causal states} of a process are the members of the range of the
function $\epsilon : \AllPasts \mapsto 2^\AllPasts$---the power set of
$\AllPasts$:
\begin{eqnarray}
\nonumber
\lefteqn{\epsilon (\past) \equiv } & & \\
\nonumber
& &   \{ \past^\prime | \Prob(\Future = \future | \Past = \past ) =
  \Prob(\Future = \future | \Past = \past^\prime ),\\
& &  {\rm for~all} \future \in \Future , \past^\prime \in \Past\} ~,
\end{eqnarray}
\label{def-of-causal-states}

that maps from histories to classes of histories.  We write the $i^{th}$ causal
state as $\CausalState_i$ and the set of all causal states as
$\CausalStateSet$; the corresponding random variable is denoted $\CausalState$
and its realization $\causalstate$.
\label{CausalStatesFunctionDefn}
\end{definition}

The cardinality of $\CausalStateSet$ is unrestricted.  $\CausalStateSet$ can be
finite, countably infinite, a continuum, a Cantor set, or something stranger
still.\footnote{Examples of all of these are given in
\citeN{Calculi-of-emergence} and \citeN{Upper-thesis}.}

We could equally well define an equivalence relation $\CausalEquivalence$ such
that two histories are equivalent iff they have the same conditional
distribution of futures, and define causal states as the equivalence classes of
$\CausalEquivalence$.  (In fact, this was the original approach of
\citeN{Inferring-stat-compl}.)  Either way, we break $\AllPasts$ into parts
that leave us in different conditions of ignorance about the
future.

Each causal state $\CausalState_i$ has a conditional distribution of futures,
$\Prob ( \Future | \CausalState_i)$.  It follows directly from
Def.~\ref{CausalStatesFunctionDefn} that no two states have the same
distribution of futures; this is not true of effective states in general.
Another immediate consequence of that definition is that
\begin{equation}
\label{conditional-distributions-of-causal-states-and-histories-are-equal}
\Prob(\Future = \future | \CausalState = \epsilon(\past)) = \Prob(\Future =
\future | \Past = \past).
\end{equation}
Again, this is not generally true of effective states.

\subsection{Causal State-to-State Transitions}

The causal state at any given time and the next value of the observed process
together determine a new causal state (Lemma \ref{automatism-lemma} below,
which doesn't rely on the following).  Thus, there is a natural relation of
succession among the causal states.

\begin{definition}
{\bf (Causal Transitions)}
The labeled transition probability $T_{ij}^{(s)}$ is the probability of
making the transition from state $\CausalState_i$ to state $\CausalState_j$ while emitting
the symbol $s \in {\cal A}$:
\begin{equation}
T_{ij}^{(s)} \equiv
  \Prob(\CausalState^\prime = \CausalState_j \ProbAnd \Future^1 = s | \CausalState = \CausalState_i ) ~,
\end{equation}
\label{CausalTransitionsDefn}
where $\CausalState$ is the current causal state and $\CausalState^\prime$
its successor on emitting $s$. We denote the set
$\{ T_{ij}^{(s)} : s \in {\cal A} \}$ by $\bf T$.
\end{definition}

\subsection{$\epsilon$-Machines}

\begin{definition}[An $\epsilon$-Machine Defined]
The {\rm $\epsilon$-machine} of a process is the ordered pair
$\{ \epsilon, {\bf T} \}$, where $\epsilon$ is the causal state function and
$\bf T$ is set of the transition matrices for the states defined by $\epsilon$.
\end{definition}

\begin{lemma}
{\bf ($\epsilon$-Machines Are Deterministic)}
For each $\CausalState_i$ and $s \in {\cal A}$, ${T}^{(s)}_{ij} > 0$ only
for that $\CausalState_j$ for which $\epsilon(\past \! s) = \CausalState_j$
iff $\epsilon(\past) = \CausalState_i$, for all pasts $\past$.
\label{automatism-lemma}
\end{lemma}

``Deterministic'' is meant in the sense of automata-theory, not dynamics.

\begin{lemma}
{\bf (Causal States Are Independent)}
The probability distributions over causal states at different times are
conditionally independent.
\label{independent-states-lemma}
\end{lemma}

This indicates that the causal states, considered as a process, define a kind
of Markov chain.  We say ``kind of'' since the class of $\epsilon$-machines is
substantially richer than the one normally associated with Markov chains
\cite{Calculi-of-emergence,Upper-thesis}.

\begin{definition}[$\epsilon$-Machine Reconstruction]
$\epsilon$-{\rm Machine reconstruction} is any procedure that given a process
$\Prob(\BiInfinity)$, or an approximation of $\Prob(\BiInfinity)$, produces
the process's $\epsilon$-machine $\{\epsilon, {\bf T} \}$.
\label{ReconstructionDefn}
\end{definition}

Given a mathematical description of a process, one can often calculate
analytically its $\epsilon$-machine. (For example, see the computational
mechanics analysis of statistical mechanical spin systems in \citeN{DNCO}.)
There are also algorithms that reconstruct $\epsilon$-machines from empirical
estimates of $\Prob(\BiInfinity)$.  Those used in \citeN{Calculi-of-emergence},
\citeN{Inferring-stat-compl}, \citeN{Hanson-thesis}, and
\citeN{Perry-Binder-finite-stat-compl}, operate in batch mode, taking the
raw data as a whole and producing the $\epsilon$-machine.  Others could work
on-line, taking in individual measurements and re-estimating the set of causal
states and their transition probabilities.

\subsection{Optimalities and Uniqueness}

\begin{theorem}
{\bf (Causal States are Maximally Prescient)}
For all $\AlternateStateSet$ and all $L \in {\Bbb Z}^{+}$,
\begin{equation}
H[\FutureL|\AlternateState] \geq H[\FutureL|\CausalState] = H[\FutureL|\Past]
~.
\end{equation}
\label{optimal-prediction-theorem}
\end{theorem}

Causal states are as good at predicting the future---are as {\it
prescient}---as complete histories.  Since the causal states can be
systematically approximated, we have shown that the upper bound on the strength
of patterns (Def.~\ref{capturing-a-pattern} and Lemma \ref{old-country-lemma})
can in fact be reached.

All subsequent results concern rival states that are as prescient as the
causal states.  We call these {\em prescient rivals} and denote a class of
them $\PrescientStateSet$.

\begin{definition}[Prescient Rivals]
{\em Prescient rivals} $\PrescientStateSet$ are states that are as
predictive as the causal states; viz., for all $L \in {\Bbb Z}^{+}$,
\begin{equation}
H[\FutureL|\PrescientState] = H[\FutureL|\CausalState] ~.
\end{equation}
\label{PrescientRivals}
\end{definition}

\begin{figure}
\epsfxsize=2.7in
\begin{center}
\leavevmode
\epsffile{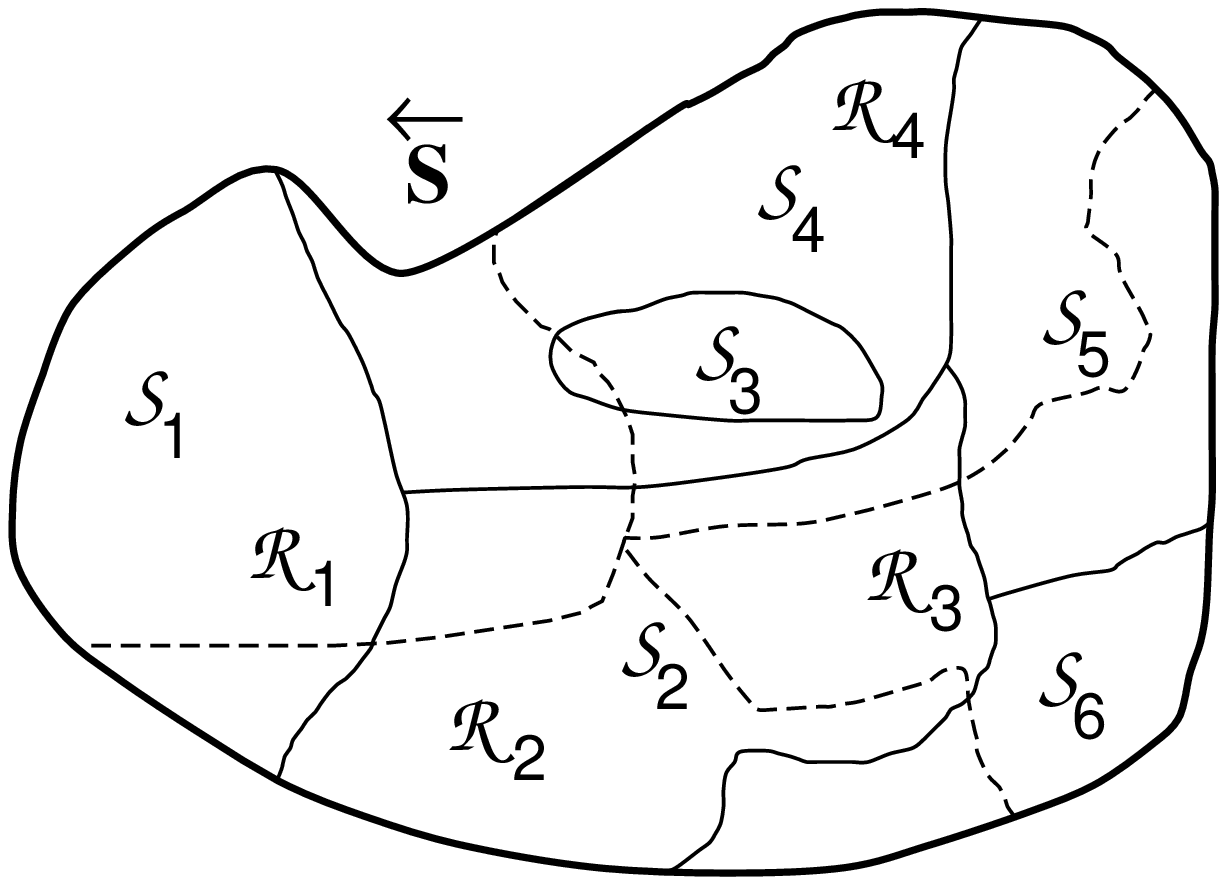}
\end{center}
\caption{An alternative class $\AlternateStateSet$ of states (delineated by
dashed lines) that partition $\AllPasts$ overlaid on the causal states
$\CausalStateSet$ (solid lines).  Here, for example, $\CausalState_2$ contains
parts of $\AlternateState_1$, $\AlternateState_2$, $\AlternateState_3$ and
$\AlternateState_4$.  Note again that the $\AlternateState_i$ need not be
compact nor simply connected, as drawn.
  }
\label{epsilon-and-bad-eta-partition}
\end{figure}

\begin{lemma}[Refinement Lemma]
For all prescient rivals $\PrescientStateSet$ and for each
$\prescientstate \in \PrescientStateSet$, there is a
$\causalstate \in \CausalStateSet$ and a measure-$0$
subset $\prescientstate_0 \subset \prescientstate$, possibly empty,
such that $\prescientstate \setminus \prescientstate_0 \subseteq \causalstate$,
where $\setminus$ is set subtraction.
\label{refinement-lemma}
\end{lemma}

\begin{figure}
\epsfxsize=2.7in
\begin{center}
\leavevmode
\epsffile{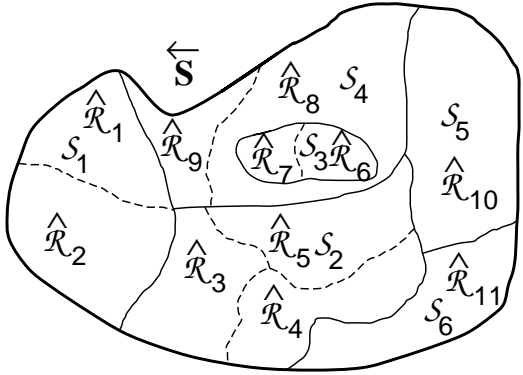}
\end{center}
\caption{A prescient rival partition $\PrescientStateSet$ must be a refinement
of the causal-state partition {\em almost everywhere}.  Almost all of each
$\PrescientState_i$ must lie within some $\CausalState_j$; the exceptions, if
any, are a set of histories of measure $0$.  Here for instance $\CausalState_2$
contains the positive-measure parts of $\PrescientState_3$,
$\PrescientState_4$, and $\PrescientState_5$.  One of these rival states, say
$\PrescientState_3$, could have member-histories in any or all of the other
causal states, if the total measure of these exceptional histories is zero.}
\label{refined-partition}
\end{figure}

The lemma becomes more intuitive if we ignore for a moment the measure-$0$ set
$\prescientstate_0$ of histories.  It then says
that any alternative partition $\PrescientStateSet$ that is as prescient as the
causal states must be a refinement of the causal-state partition.  That is,
each $\PrescientState_i$ must be a (possibly improper) subset of some
$\CausalState_j$.  Otherwise, at least one $\PrescientState_i$ would 
contain parts of at least two causal states.  Therefore, using
$\PrescientState_i$ to predict the future observables would lead to more
uncertainty about $\Future$ than using the causal states.  (Compare
Fig.~\ref{refined-partition} with Fig.~\ref{epsilon-and-bad-eta-partition}.)
Because the histories in $\prescientstate_0$ have zero probability, treating
them the ``wrong'' way makes no discernible difference to predictions.

\begin{theorem}[Causal States Are Minimal]
For all prescient rivals $\PrescientStateSet$,
\begin{equation}
\Cmu(\PrescientStateSet) \geq \Cmu(\CausalStateSet) ~.
\end{equation}
\label{minimality-theorem}
\end{theorem}

If we were trying to predict, not the whole of $\Future$ but some limited piece
$\FutureL$, the causal states might not be the simplest ones with full
predictive power.  For any value of $L$, however, the states constructed by
analogy to the causal states---the ``truncated causal states''---have
maximal prescience and minimal $\Cmu$.

The minimality theorem licenses the following definition.

\begin{definition}
{\bf (Statistical Complexity of a Process)}
The statistical complexity $\Cmu({\cal O})$ of a process $\cal O$
is that of its causal states: $\Cmu ({\cal O}) \equiv \Cmu(\CausalStateSet)$.
\label{statistical-complexity-of-a-process}
\end{definition}

\begin{theorem}
{\bf (Causal States Are Unique)}
For all prescient rivals $\PrescientStateSet$, if $\Cmu(\PrescientStateSet) =
\Cmu(\CausalStateSet)$, then there exists an invertible function between
$\PrescientStateSet$ and $\CausalStateSet$ that almost always preserves
equivalence of state: $\PrescientStateSet$ and $\eta$ are the same as
$\CausalStateSet$ and $\epsilon$, respectively, except on a set of histories of
measure $0$.
\label{uniqueness-theorem}
\end{theorem}

The remarks on Lemma \ref{refinement-lemma} also apply to the ineliminable but
immaterial measure-$0$ caveat here.

\begin{theorem}{\bf ($\epsilon$-Machines Are Minimally Stochastic)}
For all prescient rivals $\PrescientStateSet$,
\begin{equation}
H[\PrescientState^{\prime}|\PrescientState] \geq H[\CausalState^{\prime}|\CausalState] ~,
\end{equation}
where $\CausalState^{\prime}$ and $\PrescientState^{\prime}$ are the next
causal state of the process and the next $\eta$-state, respectively.
\label{minimal-stochasticity-theorem}
\end{theorem}

Finally, we relate $\Cmu$ to an information theoretic quantity that is often
used to measure complexity.

\begin{definition}[Excess Entropy]
The {\rm excess entropy} $\EE$ of a process is the mutual information between
its semi-infinite past and its semi-infinite future:
\begin{equation}
\EE \equiv I[\Future;\Past] ~.
\end{equation}
\label{ExcessEntropyDefn}
\end{definition}

Excess entropy is regularly re-introduced into the complexity-measure
literature, as ``predictive information'', ``stored information'', ``effective
measure complexity'', and so on \cite[Sec. VI]{CMPPSS}.  As these names
indicate, it is tempting to see $\EE$ as the amount of information stored in a
process (which accounts for its popularity). According to the following
theorem this temptation should be resisted.

\begin{theorem}
The statistical complexity $\Cmu$ bounds the excess entropy $\EE$:
\begin{equation}
{\EE} \leq \Cmu(\CausalStateSet) ~,
\end{equation}
with equality iff $H[\CausalState|\Future] = 0$.
\label{bounded-excess-theorem}
\end{theorem}

$\EE$ is thus only a lower bound on the true amount of information stored in
the process, namely $\Cmu(\CausalStateSet)$.

\section{Relations to Other Fields}
\label{compare-and-contrast}

\subsection{Computational and Statistical Learning Theory}

The goal of computational learning theory \cite{Kearns-Vazirani,Vapnik-nature}
is to identify algorithms that quickly, reliably, and simply lead to good
representations of a target concept, usually taken to be a dichotomy of a
feature or input space.  Particular attention is paid to ``probably approximately
correct'' (PAC) procedures \cite{Valiant-learnable}: those having a high
probability of finding a close match to the target concept among members of a
fixed representation class.  The key word here is ``fixed''. While taking the
representation class as a given is in line with implicit assumptions in most of
mathematical statistics, it seems dubious when analyzing learning in the real
world \cite{Calculi-of-emergence,Boden-precis}.

In any case, such an assumption is clearly inappropriate if our goal is pattern
discovery, and it was {\it not} made in the preceding development.  While we
plan to make every possible use of the results of computational learning theory
in $\epsilon$-machine reconstruction, we feel this theory is more properly a
part of statistical inference and, particularly, of algorithmic parameter
estimation, than of pattern discovery {\it per se}.

\subsection{Formal Language Theory and Grammatical Inference}

It is well known that formal languages can be classified into a hierarchy, the
higher levels of which have strictly greater expressive power.  The denizens of
the lowest level of the hierarchy, the regular languages, correspond to
finite-state machines and to hidden Markov models of finite dimension.  In such
cases, relatives of our minimality and uniqueness theorems are well known, and
the construction of causal states is analogous to Nerode equivalence classing
\cite{Hopcroft-Ullman}.  Our theorems, however, are {\it not} restricted to
this setting.

The problem of learning a language from observational data has been extensively
studied by linguists and computer scientists.  Unfortunately, good learning
techniques exist only for the two lowest classes in the hierarchy, the regular
and the context-free languages.  (For a good account of these procedures see
\citeN{Charniak}.)  Adapting this work to the reconstruction of
$\epsilon$-machines should be a useful area for future research.

\subsection{The Minimum Description-Length Principle}

Rissanen's {\it minimum description-length} (MDL) principle, best presented in
his book \citeyear{Rissanen-SCiSI}, is a way of picking the most concise
generative model out of a chosen family of models that are all statistically
consistent with given data. The MDL approach starts from Shannon's results
on the connection between probability distributions and codes
\cite{Shannon-and-Weaver}.

Suppose we choose a class $\cal M$ of models and are given data set $x$.  The
MDL principle tells us to use the model ${\rm M} \in {\cal M}$ that minimizes
the sum of the length of the description of $x$ given $\rm M$, plus the length
of description of $\rm M$ given $\cal M$.  The description length of $x$ is
taken to be $-\log{\Prob(x|{\rm M})}$.  The description length of ${\rm M}$ may
be regarded as either given by some coding scheme or, equivalently, by some
distribution over the members of ${\cal M}$.

Though the MDL principle was one of the inspirations of computational
mechanics, our approach to pattern discovery does not fit within Rissanen's
framework.  To mention only the most basic differences: We have no fixed class
of models ${\cal M}$; we do not use encodings of rival models or prior
distributions over them; and $\Cmu(\AlternateStateSet)$ is not a description
length.

\subsection{Connectionist Models}

Neural networks engaged in unsupervised learning \cite{Becker-unsupervised} are
often effectively doing pattern discovery.  Certainly
\citeN{Hebb-organization} had this aim in mind when proposing his learning
rule.  While such networks certainly can discover regularities and
covariations, they often represent them in ways baffling to humans.
$\epsilon$-Machines present the structures discovered by reconstruction in a
clear and distinct way, but the learning dynamics are not (currently) as well
understood as those of neural networks; see Sec.~\ref{Forwards} below.

\section{Concluding Remarks}
\label{Conclusion}

\subsection{Limitations of the Current Results}
\label{ThingsThatAreNotYetTheorems}

We made some restrictive assumptions in our development above.  Here we mention
them in order of increasing severity and consider what may be done to lift
them.

{\it 1. We know exact joint probabilities over sequence blocks of all lengths
for a process.}  The cure for this is $\epsilon$-machine reconstruction and in
the next subsection we sketch work (underway) on a statistical theory
of reconstruction.

{\it 2. The observed process takes on discrete values.}  This can probably be
addressed with only a modest cost in increased mathematical subtlety, since the
information-theoretic quantities we have used also exist for continuous
variables.  Many of our results appear to carry over to the continuous setting.

{\it 3.  The process is discrete in time.}  This looks similarly solvable,
since continuous-time stochastic process theory is moderately well developed.
It may involve sophisticated probability theory or functional analysis,
however.

{\it 4. The process is a pure time series; e.g., without spatial extent.} There
are already tricks to make spatially extended systems look like time series.
Basically, one looks at all the paths through space-time, treating each one as
if it were a time series.  While this works well for data compression
\cite{Lempel-Ziv-two-d}, it may not be satisfactory for capturing structure
\cite{DPF-thesis}.

{\it 5. The process is stationary.}  It's unclear how best to relax the
assumption of stationarity.  There are several straightforward ways of doing
so, but it is unclear how much substantive content these extensions have.  In
any case, a systematic classification of non-stationary processes is (at best)
in its infant stages.

\subsection{Directions for Future Work}
\label{Forwards}

Two broad avenues for research present themselves.

First, we have the mathematics of $\epsilon$-machines themselves.  Assumption-lifting
extensions have just been mentioned but there are many other ways to go.  One
which is especially interesting in the machine-learning context is the
trade-off between prescience and complexity.  For a given process there is a
sequence of optimal machines connecting the one-state, zero-complexity machine
with minimal prescience to the $\epsilon$-machine.  Each step on the path is
the minimal machine for a certain degree of prescience; it would be very
interesting to know what, if anything, we can say in general about the shape of
this ``prediction frontier''.

Second, there is $\epsilon$-machine reconstruction.  As we remarked
(p.~\pageref{ReconstructionDefn}), there are already several algorithms for
reconstructing machines from data.  What we need is knowledge of the {\it error
statistics} \cite{Mayo-error} of different reconstruction procedures, of the
kinds of mistakes they make and the probabilities with which they make them.
Ideally, we want to find ``confidence regions'' for the products of
reconstruction: calculating the probabilities of different degrees of
reconstruction error for a given volume of data or the amount of data needed
to be confident of a fixed bound on the error.  An analytical theory has been
developed for the expected error in reconstructing certain kinds of processes
\cite{JPC-and-Chris-Douglas}. The results are encouraging enough that
work is underway on a general theory of statistical inference for
$\epsilon$-machines---a theory analogous to what already exists in
computational learning theory and grammatical inference.

\subsection*{Acknowledgments}

This work was supported at the Santa Fe Institute under the Computation,
Dynamics, and Inference Program via NSF grant IIS-9705830 and under the
Network Dynamics Program via support from Intel Corporation.

% **************************References**************************

\bibliography{locusts}

\end{document}